\documentclass[letterpaper]{article} 
\usepackage{aaai2026}
\usepackage{times}  
\usepackage{helvet}  
\usepackage{courier}  
\usepackage[hyphens]{url}  
\usepackage{graphicx} 
\usepackage{natbib}  
\usepackage{caption} 
\usepackage{algorithm}
\usepackage{enumitem}
\usepackage{newfloat}
\usepackage{listings}
\DeclareCaptionStyle{ruled}{labelfont=normalfont,labelsep=colon,strut=off} 
\floatstyle{ruled}
\newfloat{listing}{tb}{lst}{}
\floatname{listing}{Listing}
\newcommand{\mname}{EAG-RL\xspace}
\newcommand{\newcustomfootnotesize}{\fontsize{9pt}{9pt}\selectfont}
\newcommand{\newnewcustomfootnotesize}{\fontsize{7.8pt}{7.8pt}\selectfont}

\PassOptionsToPackage{table}{xcolor}
\usepackage{xcolor}     

\usepackage[most]{tcolorbox} 
\usepackage{booktabs}
\usepackage{multirow}
\usepackage{amssymb}
\usepackage{graphicx}        
\usepackage{pifont}
\usepackage{xspace}

\usepackage{algorithm}
\usepackage{algpseudocode}
\usepackage{amsmath, amsfonts, amssymb}
\usepackage{longtable}
\usepackage{fontawesome5}
\usepackage{array}

\title{Toward Better EHR Reasoning in LLMs: Reinforcement Learning with Expert Attention Guidance}
\author{
    Yue Fang\textsuperscript{$\spadesuit$} \equalcontrib, 
    Yuxin Guo\textsuperscript{$\spadesuit$} \equalcontrib, 
    Jiaran Gao\textsuperscript{$\spadesuit$}\equalcontrib,
    Hongxin Ding\textsuperscript{$\spadesuit$}\equalcontrib,\\
    Xinke Jiang\textsuperscript{$\spadesuit$},
    Weibin Liao\textsuperscript{$\spadesuit$}, 
    Yongxin Xu\textsuperscript{$\heartsuit$},
    Yinghao Zhu\textsuperscript{$\diamondsuit$},
    Zhibang Yang\textsuperscript{$\spadesuit$},\\
    Liantao Ma\textsuperscript{$\clubsuit$}\thanks{Corresponding Author.},
    Junfeng Zhao\textsuperscript{$\spadesuit$}\footnotemark[2],
    Yasha Wang\textsuperscript{$\clubsuit$}\footnotemark[2],
}
\affiliations{
    \textsuperscript{$\spadesuit$} School of Computer Science, Peking University, Beijing, China\\
    \textsuperscript{$\heartsuit$} Key Laboratory of High Confidence Software Technologies, Ministry of Education, Beijing, China\\
    \textsuperscript{$\clubsuit$} National Engineering Research Center For Software Engineering, Peking University, Beijing, China\\
    \textsuperscript{$\diamondsuit$} School of Computing and Data Science, The University of Hong Kong, Hong Kong SAR, China\\
    \faEnvelope[regular] yuefang25@stu.pku.edu.cn\\
    
    \faGithub\ github.com/devilran6/EAG-RL
}

\usepackage{bibentry}

\begin{document}

\maketitle

\begin{abstract}
Improving large language models (LLMs) for electronic health record (EHR) reasoning is essential for enabling accurate and generalizable clinical predictions. While LLMs excel at medical text understanding, they underperform on EHR-based prediction tasks due to challenges in modeling temporally structured, high-dimensional data. Existing approaches often rely on hybrid paradigms, where LLMs serve merely as frozen prior retrievers while downstream deep learning (DL) models handle prediction, failing to improve the LLM’s intrinsic reasoning capacity and inheriting the generalization limitations of DL models. To this end, we propose \textbf{\mname}, a novel two-stage training framework designed to intrinsically enhance LLMs’ EHR reasoning ability through expert attention guidance, where expert EHR models refer to task-specific DL models trained on EHR data. Concretely, \mname first constructs high-quality, stepwise reasoning trajectories using expert-guided Monte Carlo Tree Search to effectively initialize the LLM’s policy. Then, \mname further optimizes the policy via reinforcement learning by aligning the LLM’s attention with clinically salient features identified by expert EHR models. Extensive experiments on two real-world EHR datasets show that \mname{} improves the intrinsic EHR reasoning ability of LLMs by an average of 14.62\%, while also enhancing robustness to feature perturbations and generalization to unseen clinical domains. These results demonstrate the practical potential of \mname{} for real-world deployment in clinical prediction tasks. Our code have
been available at https://github.com/devilran6/EAG-RL.
\end{abstract}



\section{Introduction}





Large Language Models (LLMs) have shown strong capabilities across a wide spectrum of unstructured medical text processing tasks such as clinical note classification and report summarization~\cite{jahan2024comprehensive,chen2023large,zhou2023survey}, showing potential to assist physicians in accurate diagnosis~\cite{Wang_Zhao_Ouyang_Wang_Shen_Segmentor,Xu_Xu_Wang_Liu_Zhu_Mcauley_Diego,liu2024large} and treatment planning~\cite{Jiang_Zhang_Xu_Qiu_Fang_Wang_Tang_Ding_Chu_Zhao_et}. 
While effective in above tasks, LLMs still underperform on clinical prediction tasks based on \textbf{E}lectronic \textbf{H}ealth \textbf{R}ecords (EHR)~\cite{brown2024not,chen2024clinicalbench}, which contain time-series values prevalent in healthcare systems and convey crucial physiological signals of patient health.
\begin{figure}[t]
    \centering
    \includegraphics[scale=0.14]{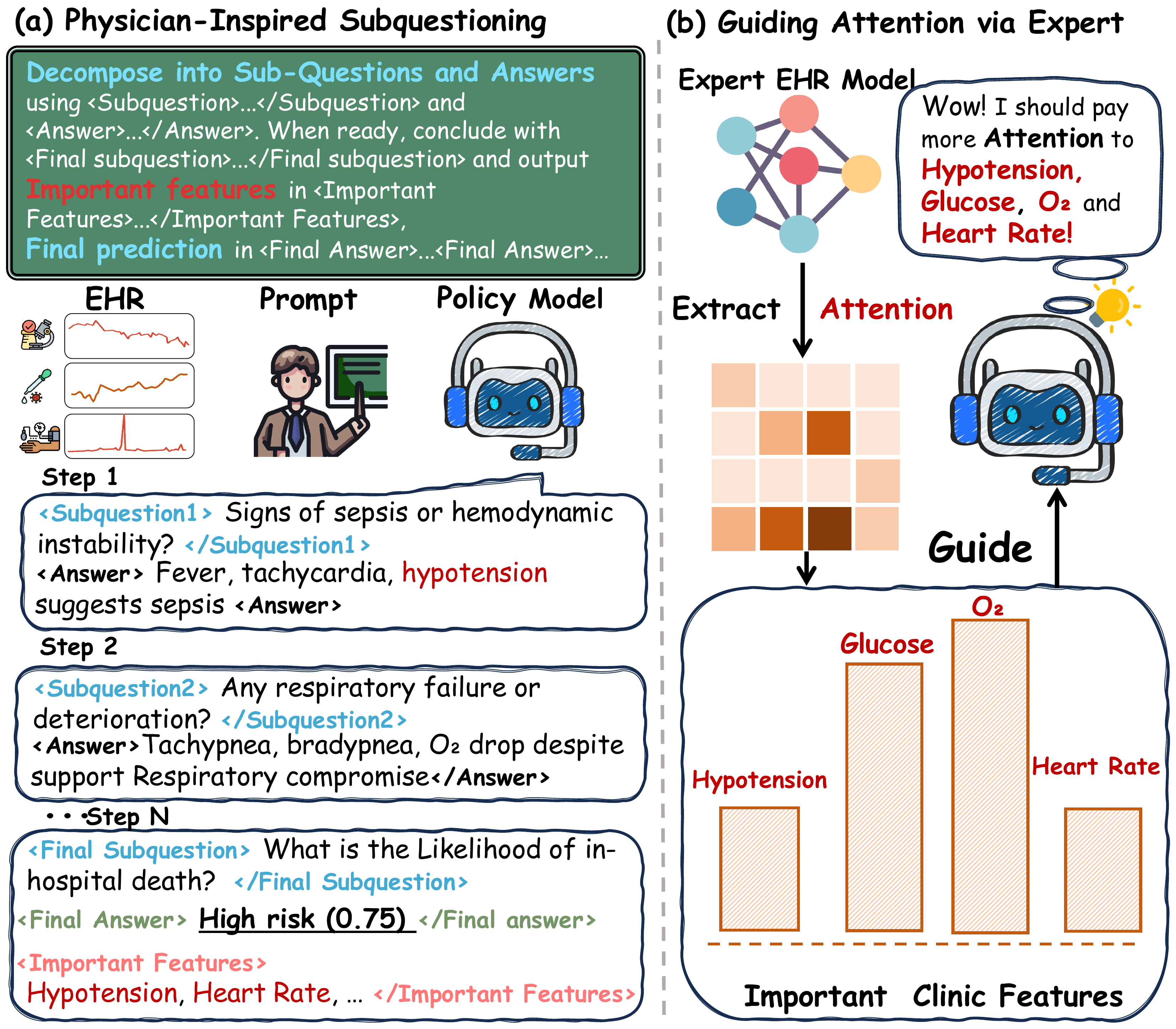}
    \caption{
(a) Physician-inspired subquestioning, 
(b) Guiding attention via expert.}
    \label{fig:introduction}
\end{figure}
Recent studies~\cite{brown2024not,chen2024clinicalbench,zhu2024clinicrealm,wang2023can} show that conventional deep learning models, referred to as \textit{expert EHR models} due to their task-optimized architectures, remain the dominant choice for clinical prediction. Although significantly outperforming LLMs on EHR tasks~\cite{zhu2024clinicrealm}, their reliance on fixed input schemas renders them brittle to variations in feature order, availability, and encoding, which are common across institutions with heterogeneous EHR systems and ultimately limit their generalizability. LLMs, by contrast, exhibit stronger generalization and hold promise as unified reasoning engines capable of learning transferable paradigms for robustly interpreting heterogeneous EHR data. Yet most existing approaches~\cite{jiang2023graphcare,xu2025dearllm,xu2024ram,nguyen2024carer,zhu2024emerge} adopt hybrid, tool-using paradigms, where LLMs serve primarily as static prior retrievers, while downstream expert models handle final prediction, failing to improve the LLM’s intrinsic reasoning capacity and inheriting traditional models’ generalization limits. This gap highlights the urgent need to \textbf{intrinsically strengthen LLMs’ EHR reasoning capacity} for clinical decision-making.

Recently, \textbf{R}einforcement \textbf{L}earning (RL)~\cite{sutton1999reinforcement,kaelbling1996reinforcement} has emerged as a powerful paradigm to enhance reasoning capacity, with models like OpenAI-O1~\cite{openaio1} and DeepSeek-R1~\cite{guo2025deepseekr1} achieving strong performance through learned stepwise reasoning policies. Inspired by these advances, we ask: \textit{Can RL similarly benefit LLMs in EHR-based clinical prediction tasks?} Notably, real-world clinical prediction reflects a \textit{hypothetico-deductive reasoning} process~\cite{norman2010diagnostic}, where physicians iteratively pose diagnostic subquestions and integrate evidence through stepwise reasoning to reach a well-supported conclusion. This leads to our first motivation(Figure~\ref{fig:introduction}(a)): 
\textbf{M\#1.} \textit{To enhance LLMs’ EHR reasoning capacity via RL by imitating physicians’ hypothesis refinement through stepwise subquestioning and evidence integration.}

However, the high-dimensional and temporally evolving nature of EHR demands reasoning policies that can dynamically attend to clinically salient features, similar to how physicians iteratively analyze important features by assessing their trends and clinical implications. Yet, sparse outcome rewards provide limited guidance for learning such fine-grained attention behaviors. Interestingly, prior works~\cite{ma2020concare,xu2023seqcare,xu2023vecocare} show that transformer-based expert EHR models can capture clinically salient features through attention mechanisms. This insight motivates us to distill the attention patterns of expert EHR models as auxiliary reward signals, guiding LLMs to assign greater focus to clinically salient features during training. This leads to our second motivation(Figure~\ref{fig:introduction}(b)):
\textbf{M\#2.} \textit{To distill attention from expert EHR models to guide LLMs in attending to clinically salient features.}

This dual motivations give rise to our central goal: \textbf{to enhance LLMs’ EHR reasoning capabilities via expert-attention guided RL}. However, realizing this goal remains unexplored and faces substantial challenges:

\noindent\textbf{\ding{182} \textbf{Challenge\#1: How to construct high-quality stepwise trajectories for effective policy initialization?}} RL on reasoning tasks often suffers from unstable convergence and poor sample efficiency when initialized from a weak policy~\cite{wang2025beyond,xu2025kdrl}. Prior work demonstrates that strong initialization via \textbf{S}upervised \textbf{F}ine-\textbf{T}uning(SFT) can substantially improve RL efficiency~\cite{chu2025sft,chen2025towards,wang2025beyond}. However, the scarcity of real-world multi-step SFT data showing how clinicians reason over patient records  poses a major obstacle. The challenge lies in constructing high-quality, stepwise trajectories that capture realistic clinical reasoning patterns for effective policy initialization.

\noindent\textbf{\ding{183} Challenge\#2. How to extract reliable supervision from expert attention to guide EHR reasoning?}
Although expert EHR models can highlight clinically salient features via attention~\cite{yang2023attention,ma2020concare,xu2023seqcare,xu2023vecocare}, directly aligning LLM attention with expert models is nontrivial due to semantic and architectural mismatches~\cite{hao2023one}, which risks introducing spurious supervision and leading to suboptimal policy updates. Additionally, extracting attention from LLMs imposes prohibitive computational overhead~\cite{dao2022flashattention}, making it impractical for frequent RL sampling. This calls for a scalable and semantically coherent alignment strategy that bridges model discrepancies while preserving training efficiency.

\noindent\textbf{\ding{184} Challenge\#3. How to encourage exploration of informative clinical patterns?} Recent studies indicate that RL algorithms like GRPO~\cite{shao2024deepseekmath} are prone to entropy collapse, which leads to converging prematurely on low-entropy reasoning trajectories while failing to explore diverse, high-entropy alternatives, trapping in local optima~\cite{yu2025dapo}. In EHR reasoning, LLMs risk converging on attending to a narrow set of high-confidence clinical features while overlooking truly critical ones. The challenge lies in designing reward mechanisms that adaptively amplify the influence of under-attended yet clinically meaningful features, thereby encouraging the model to explore high-entropy but clinical meaningful reasoning patterns.


To address these challenges, we propose \underline{\textbf{E}}xpert-\underline{\textbf{A}}ttention \underline{\textbf{G}}uided \underline{\textbf{RL}} (\textbf{\mname}), a two-stage framework designed to intrinsically strengthen LLMs’ EHR reasoning capacity via expert-guided policy optimization. For \textbf{Challenge\#1}, we introduce \textit{Expert-Guided Trajectory Distillation}, which constructs high-quality, stepwise reasoning trajectories via \textit{Expert-guided Monte Carlo Tree Search}, enabling effective policy initialization.
For \textbf{Challenge\#2}, we introduce \textit{Attention-Aligned Policy Optimization}, an RL-based stage that leverages expert attention as auxiliary reward through a lightweight alignment strategy, which quantifies the overlap between clinically salient features identified by the LLM and expert model via Jaccard similarity.
For \textbf{Challenge\#3}, we introduce \textit{Entropy-Aware Adaptive Up Clipping}, which adaptively adjusts the clipping bound based on the entropy of salient features within reasoning trajectories, encouraging exploration of high-entropy yet meaningful reasoning patterns. In summary, our contributions are as follows:
\begin{itemize}[leftmargin=*,noitemsep,topsep=2pt]
    \item \textbf{Insightfully}, we demonstrate that expert EHR models can serve as effective policy supervisors for LLM, and propose EAG-RL as the first framework leveraging expert attention for enhancing LLMs’ EHR reasoning capacity.

    \item \textbf{Technically}, we design a novel two-stage optimization framework: (1) a \textit{warm-up stage} utilizing expert-guided MCTS to distill high-quality reasoning trajectories for effective policy initialization, and (2) a \textit{reinforcement stage}  integrating expert attention alignment and entropy-aware adaptive up clipping to to encourage exploration of clinically meaningful reasoning patterns.
    \item \textbf{Experimentally}, we conduct extensive experiments to validate EAG-RL on two real-world EHR datasets, demonstrating superior performance over state-of-the-art baselines. Further ablation and analysis substantiate the reasonableness and generalizability of EAG-RL.  
\end{itemize}




\begin{figure*}[ht]
	\centering{\includegraphics[width=18cm]{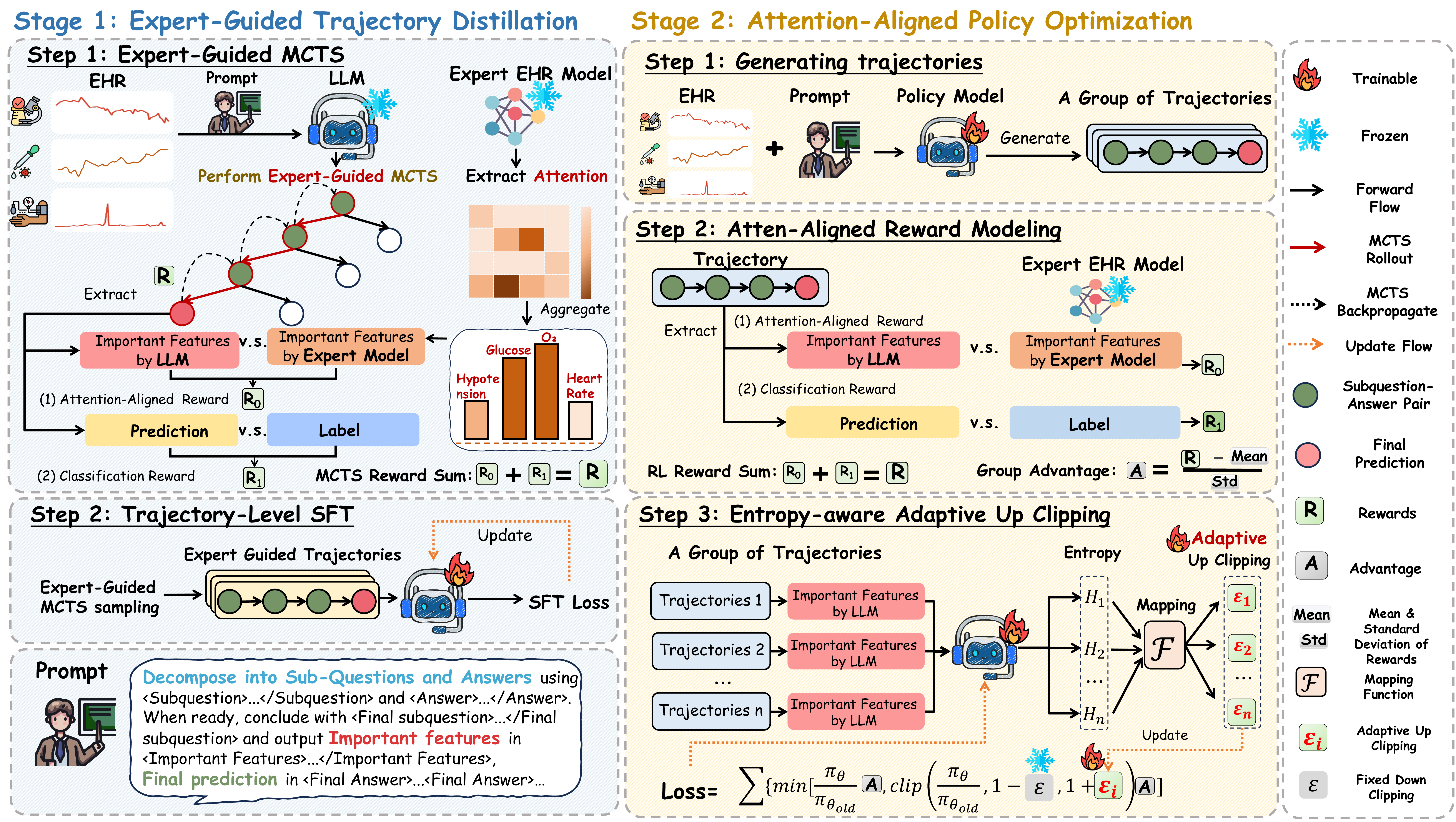}}
 	\caption{Illustration of \mname.} 
	\label{fig:modelStructure}
\end{figure*}

\section{Task Definition}


\noindent\textbf{Definition 1 (EHR Dataset).}  
A patient's EHR is represented as a sequence of $T$ time-ordered visits $\mathbf{X} = [\mathbf{x}_1, \mathbf{x}_2, \cdots, \mathbf{x}_T]$, where each visit $\mathbf{x}_t = \{l_{t,1}, l_{t,2}, \cdots, l_{t,n_t}\}$ contains $n_t$ lab test features.

\noindent\textbf{Definition 2 (Mortality Prediction).}  
Given $\mathbf{X}$, predict whether the patient will survive the hospital stay. The label $y \in \{0,1\}$ denotes death ($y = 1$) or survival ($y = 0$).

\noindent\textbf{Definition 3 (Readmission Prediction).}  
Given $\mathbf{X}$, predict whether the patient will be readmitted within 30 days after discharge. The label $y \in \{0,1\}$ indicates readmission ($y = 1$) or not ($y = 0$).



\section{Methodology}

\subsection{Overview}


As illustrated in Figure~\ref{fig:modelStructure}, \mname{} includes two stages: 
\begin{itemize}[leftmargin=*,noitemsep,topsep=2pt]
    \item \textbf{\# Stage 1: Expert-Guided Trajectory Distillation}  constructs high quality, step-wise expert-guided reasoning trajectories to effectively initialize the LLM’s policy.
    \item \textbf{\# Stage 2: Attention-Aligned Policy Optimization} further optimizes LLM’s policy via RL, guided by attention alignment with the \textit{expert EHR model}. 
\end{itemize}

\subsection{Expert-Guided Trajectory Distillation}
This stage aims to initialize the LLM’s EHR reasoning ability through expert-guided, high-quality reasoning trajectories. To achieve this, we first generate multi-step trajectories using \textit{Expert-Guided Monte Carlo Tree Search}, followed by \textit{Trajectory-Level Supervised Fine-Tuning} to instill clinical reasoning patterns into the model.

\noindent\textbf{Prompt-based Question Decomposition.} 
Clinical reasoning often follows a hypothetico-deductive process~\cite{norman2010diagnostic}, where clinicians iteratively pose subquestions to refine diagnostic hypotheses. Inspired by this, we design prompt \( \mathcal{P}_{\text{QD}}(\mathbf{X}) \) to guide LLMs in decomposing complex EHR tasks into a sequence of subquestions and intermediate answers, simulating step-by-step clinical reasoning (Figure~\ref{fig:introduction}(a)). Each step is represented using standardized tags: \texttt{<Subquestion>} and \texttt{<Answer>}, forming a structured reasoning unit. Once the model determines sufficient evidence has been gathered, it generates a final prediction using \texttt{<Final subquestion>}, \texttt{<Important Features>}, and \texttt{<Final answer>}. Full prompt templates are provided in Appendix D. A reasoning trajectory is defined as an ordered sequence of subquestion-answer pairs, $\tau=\{(q_1, a_1), (q_2, a_2), \dots, (q_T, a_T)\}$, where each $a_i$ is either an intermediate reasoning output $\mathcal{S}(q_i)$ for $i<T$, or a final prediction $(y, \mathcal{C})$ at step $i=T$. Here, $y$ denotes the predicted outcome, and $\mathcal{C} = \{c_1, c_2, \dots, c_K\}$ is a set of salient clinical features explicitly identified by the LLM as salient evidence in support of the predicted outcome.

\noindent\textbf{Expert-Guided Monte Carlo Tree Search.} 
To construct high-quality reasoning trajectories aligned with clinical logic, we adopt Monte Carlo Tree Search (MCTS)~\cite{kocsis2006bandit, coulom2006efficient}, which explores the space of subquestion-answer paths defined by \( \mathcal{P}_{\text{QD}} \) and efficiently balances exploration and exploitation to identify high-reward inference traces. To guide exploration toward salient clinical features, we further incorporate attention signals from a pretrained \textit{expert EHR model} \( \mathcal{M}_{\text{exp}} \), instantiated as Concare~\cite{ma2020concare}, a lightweight Transformer-based model specifically designed for EHR prediction and capable of capturing clinically salient features. The expert model's attention serves as an interpretable proxy for clinical relevance, steering MCTS toward salient clinical features and diagnostically meaningful subquestions. Specifically, \textit{Expert-Guided MCTS} iteratively constructs a reasoning tree, where each node represents a reasoning state \( s \in \mathcal{S} \) defined by a partial trajectory \( \tau_{1:t} = \{(q_1, a_1), \dots, (q_t, a_t)\} \). An edge corresponds to adding a new subquestion-answer pair, extending the trajectory to \( \tau_{1:t+1} \). At each state \( s \), the action space \( \mathcal{A}(s) \) consists of candidate subquestions \( q_{t+1} \) generated by the LLM, conditioned on the current trajectory. The goal is to discover a complete trajectory \( \tau_{1:T} \) that yields an accurate and clinically meaningful prediction. More details and pseudo-code are given in Appendix C.
\begin{itemize}[leftmargin=1.5em]
    \item \textbf{Selection.} At each iteration, we select a leaf node for expansion by traversing the tree using the Upper Confidence Bound (UCT) criterion~\cite{kocsis2006bandit}, which balances exploitation and exploration. For a node \( s \) and action \( a \in \mathcal{A}(s) \), the next subquestion is chosen by maximizing:
    \begin{equation}
    \newnewcustomfootnotesize
    \mathrm{UCT}(s, a) = Q(s, a) + \lambda \cdot \sqrt{\frac{\log N(s)}{N(s, a)}},
    \end{equation}
    where \( Q(s, a) \) is the estimated action value, \( N(s) \) is the visit count of node \( s \), \( N(s, a) \) is the count for action \( a \), and \( \lambda \) controls exploration strength.

    \item \textbf{Expansion.} If the selected node is non-terminal, we generate \( d \) candidate subquestions \( \{q_{t+1}^{(j)}\}_{j=1}^d \), each conditioned on the current trajectory \( \tau_{1:t} \). For each \( q_{t+1}^{(j)} \), the LLM produces an answer \( a_{t+1}^{(j)} \), which is either an intermediate reasoning output or a final prediction, based on a termination indicator \( \delta^{(j)} \in \{0, 1\} \):
    \begin{equation}
    \newnewcustomfootnotesize
    a_{t+1}^{(j)} =
    \begin{cases}
    \mathcal{S}(q_{t+1}^{(j)}), & \text{if } \delta^{(j)} = 1, \\
    (y^{(j)}, \mathcal{C}^{(j)}), & \text{if } \delta^{(j)} = 0,
    \end{cases}
    \label{eq:expansion_answer}
    \end{equation}
    Each resulting pair \( (q_{t+1}^{(j)}, a_{t+1}^{(j)}) \) extends the trajectory and is added as a new child node. If the selected node is terminal, expansion is skipped.

    \item \textbf{Simulation.} To estimate the expected future reward of an expanded node efficiency, we perform a single-step rollout, recursively extending the trajectory until reaching a terminal state. At each step, the LLM generates \( d \) candidate subquestions and selects the most helpful one based on a local reward signal, following~\cite{hao2023reasoning} to reduce noise and cost. The local reward evaluates the usefulness of a candidate \( q_{t+1} \) under the current prefix \( \tau_{1:t} \), using a dedicated evaluation prompt \( \mathcal{P}_{\text{H}} \) (see Appendix~D):
    \begin{equation}
    \newnewcustomfootnotesize
    r(s, a) = \mathcal{M}_{\mathcal{P}_{\text{H}}}\left(\tau_{1:t} \cup \{(q_{t+1}, \varnothing)\}\right),
    \label{eq:rollout_reward}
    \end{equation}
    where \( \mathcal{M} \) denotes the LLM and \( \varnothing \) indicates the subquestion is scored without an answer. This encourages exploration of informative, clinically relevant reasoning paths.

    \item \textbf{Backpropagation:} After reaching a terminal node, Expert-Guided MCTS backpropagates the cumulative reward to update all visited nodes. We integrate two complementary reward signals:
    
    (1) \textit{Classification reward} \( \mathcal{R}_{\mathrm{cls}} \in \mathbb{R} \) captures the model’s prediction confidence and its directional margin. Let \( \hat{y} \in (0,1) \) denote the predicted probability, \( y^\star \in \{0,1\} \) the ground-truth label, and \( \theta = 0.5 \) the decision threshold:
    \begin{equation}
    \newnewcustomfootnotesize
    \mathcal{R}_{\mathrm{cls}} = \log\left( y^\star \cdot \hat{y} + (1 - y^\star)(1 - \hat{y}) \right) + \Delta,
    \label{eq:cls_reward}
    \end{equation}
    where \( \Delta \) is a margin-aware bonus:
    \begin{equation}
    \newnewcustomfootnotesize
    \Delta =
    \begin{cases}
    \beta(\hat{y} - \theta), & \text{if } y^\star = 1 \land \hat{y} > \theta, \\
    \beta(\theta - \hat{y}), & \text{if } y^\star = 0 \land \hat{y} < \theta, \\
    0, & \text{otherwise}.
    \end{cases}
    \label{eq:soft_outcome_reward}
    \end{equation}

    (2) \textit{Attention alignment reward} \( \mathcal{R}_{\mathrm{att}} \in [0,1] \) provides a semantically coherent and computationally lightweight strategy to guide the model’s attention toward clinically meaningful information.
    Formally, \( \mathcal{R}_{\mathrm{att}} \) is computed as the Jaccard similarity between the model-extracted features \( \mathcal{C} \) and the expert-highlighted features \( \mathcal{C}_\mathrm{exp} \):
    \begin{equation}
    \newnewcustomfootnotesize
    \mathcal{R}_{\mathrm{att}} = \frac{|\mathcal{C} \cap \mathcal{C}_\mathrm{exp}|}{|\mathcal{C} \cup \mathcal{C}_\mathrm{exp}|}.
    \label{eq:attn_alignment}
    \end{equation}
    This reward encourages alignment with expert insight while maintaining training efficiency.
   
    The final reward is computed as a convex combination:
    \begin{equation}
    \newnewcustomfootnotesize
    \mathcal{R} = \lambda \cdot \mathcal{R}_{\mathrm{cls}} + (1 - \lambda) \cdot \mathcal{R}_{\mathrm{att}},
    \label{eq:total_reward}
    \end{equation}
    where \( \lambda \in [0,1] \) balances predictive accuracy and clinical alignment. The reward \( \mathcal{R} \) is backpropagated along the search path to promote trajectories that are both outcome-correct and clinically meaningful.
\end{itemize}

\noindent\textbf{Trajectory-Level Supervised Fine-Tuning.} 
\label{sec:trajectory_sft}
To distill high-quality reasoning behavior, we construct a SFT dataset \( \mathcal{D}_{\text{SFT}} \) by selecting the top-\(k\) reward trajectories from \textit{Expert-Guided MCTS}. Each trajectory \( \tau \) represents a complete step-by-step reasoning path. Given the original clinical question \( \mathcal{P}_{\text{QD}}(\mathbf{X}) \), the model \( \mathcal{M} \) is trained to generate the entire trajectory:
\begin{equation}
\newnewcustomfootnotesize
\mathcal{L}_{\text{SFT}} = - \sum_{\tau \in \mathcal{D}_{\text{SFT}}} \log \mathcal{M}(\tau \mid  \mathcal{P}_{\text{QD}}(\mathbf{X}) ),
\label{eq:sft_loss_seq}
\end{equation}

\subsection{Attention-Aligned Policy Optimization}
This stage aims to further optimize the LLM’s reasoning policy through RL, guided by expert model's attention. To achieve this, an \textit{Attention-Aligned Reward} is introduced to  encourage the LLM to focus on meaningful features identified by the expert, and an \textit{Entropy-Aware Adaptive Up Clipping} mechanism is applied to amplify learning signals for high-entropy reasoning paths, promoting exploration of uncertain but potentially informative clinical features.

\noindent\textbf{Attention-Aligned Reward Modeling.} 
\label{sec:attention_aligned_reward}
 To supervise policy optimization, we employ a composite reward that encourages both accurate predictions and clinically grounded reasoning. Specifically, we combine the outcome-based reward \( \mathcal{R}_{\mathrm{cls}} \), reflecting prediction confidence and correctness (Eq.~\ref{eq:cls_reward}), and the attention alignment reward \( \mathcal{R}_{\mathrm{att}} \), quantifying consistency with expert-highlighted features from \( \mathcal{M}_{\text{exp}} \) (Eq.~\ref{eq:attn_alignment}). The total reward \( \mathcal{R} \) is computed as a convex combination (Eq.~\ref{eq:total_reward}) and provides fine-grained feedback to guide clinically meaningful trajectory refinement.

\noindent\textbf{Entropy-Aware Adaptive Up Clipping.} 
\label{sec:entropy_clipping}
While DAPO partially mitigates entropy collapse by decoupling the clipping mechanism to encourage exploration~\cite{yu2025dapo} , it still applies static upper and lower bounds across all trajectories, lacking adaptation to individual uncertainty. Consequently, it fails to strengthen learning signals for trajectories that are rare yet clinically informative. To address this, we introduce  \textit{Entropy-Aware Adaptive Up Clipping} to adaptively adjust the clipping bound based on token-level entropy over model-attended clinical features. This adaptive strategy enables trajectory-specific modulation of update strength, amplifying learning signals for uncertain, promising paths while constraining updates for overconfident ones.

For each trajectory \( \tau \), let \( \mathcal{C} \) denote the set of key clinical tokens. We compute the average predictive entropy:
\begin{equation}
\newcustomfootnotesize
\bar{H}(\tau) = \frac{1}{|\mathcal{C}|} \sum_{t \in \mathcal{C}} H_t, \quad
H_t = - \sum_{v \in \mathcal{V}} p_t(v) \log p_t(v),
\label{eq:avg_entropy_c}
\end{equation}
where \( \mathcal{V} \) is the vocabulary and \( p_t(v) \) the predictive distribution at token \( t \). Then, we map \( \bar{H}(\tau) \) to a trajectory-specific clipping bound \( \epsilon(\tau) \in [\epsilon_{\min}, \epsilon_{\max}] \) using min-max normalization across all \( g \) sampled trajectories:
\begin{equation}
\newcustomfootnotesize
\epsilon(\tau) = \epsilon_{\min} + (\epsilon_{\max} - \epsilon_{\min}) \cdot \frac{\bar{H}(\tau) - H_{\min}}{H_{\max} - H_{\min}},
\label{eq:clip_mapping}
\end{equation}
where \( H_{\min} \) and \( H_{\max} \) denote the minimum and maximum entropy values in the group. We use \( \epsilon_{\min} = 0.2 \), \( \epsilon_{\max} = 0.4 \) in our implementation.

\noindent\textbf{RL Training.} Building on the reward design and entropy-adaptive clipping strategy, we optimize the model policy using a GRPO-style objective~\cite{shao2024deepseekmath}, further enhanced with trajectory-level adaptive clipping. Following DAPO~\cite{yu2025dapo}, we adopt an asymmetric clipping scheme. However, instead of using static bounds, we introduce a trajectory-specific adaptive upper bound \( \epsilon(\tau) \) (defined in Eq.~\ref{eq:clip_mapping}). The clipping function is:
\begin{equation}
\newcustomfootnotesize
\phi(r_t; \varepsilon, \epsilon(\tau)) = \max\left(1 - \varepsilon,\ \min(r_t,\ 1 + \epsilon(\tau)) \right),
\label{eq:clipping_function}
\end{equation}
where \( \varepsilon \) is a fixed lower bound and \( r_t \) is the token-level importance ratio. Based on this, the optimization objective is:
{\newnewcustomfootnotesize
\begin{equation}
\mathcal{J}(\theta) = \mathbb{E}_{\tau \sim \pi_{\theta_{\text{old}}}} \left[ \sum_{t=1}^{|\tau|} \min\left( r_t \cdot \hat{A},\ \phi(r_t; \varepsilon, \epsilon(\tau)) \cdot \hat{A} \right) \right],
\label{eq:grpo_loss_phi}
\end{equation}
}
where \( \hat{A} \) denotes the group-normalized advantage (see Appendix~C).

\begin{table*}[!ht]
\centering
\fontsize{9pt}{9pt}\selectfont
\setlength{\tabcolsep}{6pt}
\renewcommand{\arraystretch}{1}

\begin{tabular}{c|l|cc|cc|cc}
\toprule
\rowcolor{gray!10}
\multicolumn{2}{c|}{\hspace{-35pt}\textbf{Method}} 
& \multicolumn{2}{c|}{\textbf{TJH Mortality}} 
& \multicolumn{2}{c|}{\textbf{MIMIC-IV Mortality}} 
& \multicolumn{2}{c}{\textbf{MIMIC-IV Readmission}} \\
\rowcolor{gray!10}
Category & Variant 
& AUROC ($\uparrow$) & AUPRC ($\uparrow$) 
& AUROC ($\uparrow$) & AUPRC ($\uparrow$) 
& AUROC ($\uparrow$) & AUPRC ($\uparrow$) \\
\midrule

\multicolumn{8}{c}{\textit{Qwen2.5-7B-Instruct}} \\
\midrule
\multirow{3}{*}{Prompt-based} 
& Vanilla & 71.99±2.78 & 64.27±4.44 & 53.90±2.74 & 10.68±2.58 & 57.76±4.29 & 26.53±3.91 \\
& Think \& Answer & 79.83±2.68 & 70.87±4.61 & 61.57±7.12 & 13.58±3.17 & 55.86±3.98 & 25.32±4.02 \\
& Question Decomposition & 81.66±2.85 & 73.26±4.53 & 63.49±6.14 & 15.08±4.48 & 59.29±3.85 & 26.26±3.68 \\
\midrule
\multirow{2}{*}{SFT} 
& SFT & 81.20±3.06 & 75.14±4.39 & 64.70±6.63 & 16.99±5.96 & 51.93±4.81 & 23.25±3.84 \\
& EAG-RL(Stage-1) & 83.64±2.75 & 77.60±4.16 & \underline{69.33±5.74} & 17.90±5.20 & \underline{60.00±4.67} & \underline{28.00±4.47} \\
\midrule
\multirow{3}{*}{RL} 
& EAG-RL(Stage-1)+GRPO & \underline{85.82±2.72} & \underline{78.52±4.40} & 65.78±6.62 & 19.53±6.87 & 53.67±4.62 & 26.03±4.39 \\
& EAG-RL(Stage-1)+DAPO & 85.67±2.62 & 77.88±4.38 & 73.93±5.18 & \underline{19.71±5.71} & 57.80±4.25 & 26.74±3.94 \\
& \textbf{EAG-RL(Stage-1+Stage-2)} & \textbf{87.70±2.49} & \textbf{82.95±3.64} & \textbf{77.21±5.04} & \textbf{23.99±6.97} & \textbf{61.09±4.48} & \textbf{29.92±0.46} \\
\midrule

\multicolumn{8}{c}{\textit{LLaMA3.1-8B-Instruct}} \\
\midrule
\multirow{3}{*}{Prompt-based}
& Vanilla & 71.98±3.58 & 67.75±4.59 & 58.73±6.33 & 11.77±3.26 & 52.88±5.06 & 25.75±4.36 \\
& Think \& Answer & 57.34±3.75 & 50.26±4.18 & \underline{60.79±6.29} & 10.33±2.99 & 55.41±6.51 & 10.85±2.91 \\
& Question Decomposition &74.15±3.41  &64.88±4.83  & 58.88±6.84 & 12.71±3.70 & 54.90±5.18 & \underline{27.06±4.53} \\
\midrule
\multirow{2}{*}{SFT} 
& SFT & 76.45±3.36 & 67.84±4.91 & 47.24±7.69 & 9.79±3.19 &52.49±4.85  &24.11±3.92  \\
& EAG-RL(Stage-1) & \underline{80.67±3.03} & \underline{74.39±4.37} & 57.31±7.39 & 13.16±3.83 & \underline{55.45±3.87} & 27.04±4.44 \\
\midrule
\multirow{3}{*}{RL} 
& EAG-RL(Stage-1)+GRPO & 80.49±3.08 & 72.17±4.75 & 56.82±4.37 & \textbf{15.64±5.80} & 52.98±4.24 & 25.52±3.94 \\
& EAG-RL(Stage-1)+DAPO & 78.89±3.20 & 72.19±4.40 & 59.26±6.43 & \underline{13.82±4.08} & 55.12±3.99 & 26.83±4.30 \\
& \textbf{EAG-RL(Stage-1+Stage-2)} & \textbf{84.34±2.71} & \textbf{76.36±4.64} & \textbf{62.17±5.26} & 12.51±3.00 & \textbf{55.65±4.28} & \textbf{27.99±4.51} \\
\midrule

\multicolumn{8}{c}{\textit{Qwen2.5-3B-Instruct}} \\
\midrule
\multirow{3}{*}{Prompt-based}
& Vanilla & 66.27±2.74 & 57.63±4.11 & 51.10±6.33 & 10.06±3.41 & 54.76±3.79 & 26.27±4.19 \\
& Think \& Answer & 59.35±4.04 & 53.42±4.88 & 54.88±5.88 & 10.88±2.74 & 57.92±4.85 & 27.09±4.58 \\
& Question Decomposition & 70.97±3.34 & 59.07±4.52 & 57.88±7.78 & 12.69±5.60 & 56.82±4.96 & 28.35±5.29 \\
\midrule
\multirow{2}{*}{SFT} 
& SFT & 69.01±3.24 & 59.55±4.60 & 55.60±6.38 & 13.47±5.27 & 51.84±4.72 & 24.83±4.39 \\
& EAG-RL(Stage-1) & 77.49±6.88 & 67.91±10.08 & 67.87±4.95 & 14.62±4.04 & \underline{61.15±4.46} & \underline{29.62±4.75} \\
\midrule
\multirow{3}{*}{RL} 
& EAG-RL(Stage-1)+GRPO & 75.78±5.76
& 68.70±8.11 & \underline{67.95±5.87} & \underline{16.18±4.45} & 56.23±4.53 & 26.17±3.83 \\
& EAG-RL(Stage-1)+DAPO & \underline{79.33±4.90} & \underline{72.32±7.30} & 65.64±6.43 & 15.42±4.57 & 57.68±4.83 & 28.14±4.32 \\
& \textbf{EAG-RL(Stage-1+Stage-2)} & \textbf{80.31±4.89} & \textbf{73.46±7.48} & \textbf{70.40±5.46} & \textbf{17.95±0.51} &\textbf{61.23±4.91}  &\textbf{31.76±5.21}  \\
\bottomrule
\end{tabular}

\caption{\textbf{Performance comparison on TJH and MIMIC-IV datasets.}
We report AUROC and AUPRC (\%) for each task. 
\textbf{Bold} indicates the best-performing method, and \underline{underline} denotes the second-best across all methods.}
\label{tab:stage_results}
\end{table*}

\begin{table}[ht]
\centering
\fontsize{9pt}{9pt}\selectfont
\setlength{\tabcolsep}{3pt}
\renewcommand{\arraystretch}{1.1}

\begin{tabular}{l|cc|cc}
\toprule
\rowcolor{gray!10}
\multicolumn{1}{l|}{\textbf{Method}} 
& \multicolumn{2}{c|}{\textbf{TJH Mortality}} 
& \multicolumn{2}{c}{\textbf{MIMIC-IV Mortality}} \\
\rowcolor{gray!10}
& AUROC ($\uparrow$) & AUPRC ($\uparrow$) 
& AUROC ($\uparrow$) & AUPRC ($\uparrow$) \\
\midrule
\textbf{EAG-RL} & \textbf{87.70±2.49} & \textbf{82.95±3.64} & \textbf{77.21±5.04} & \textbf{23.99±6.97} \\
\midrule
\textit{w/o Stage-1} & 85.34±2.58 & 79.03±4.06 & 76.23±4.26 & 20.83±5.85 \\
\textit{w/o Stage-2} & 83.64±2.75 & 77.60±4.16 & 69.33±5.74 & 17.90±5.20 \\
\textit{w/o} $\mathcal{R}_{\mathrm{att}}$ & 85.20±2.62 & 76.64±4.40 & 75.58±4.28  & 20.32±5.61  \\
\textit{w/o} $\epsilon(\tau)$ & 80.84±2.96 & 72.71±4.53 & 66.24±5.12  & 14.82±4.03  \\
\bottomrule
\end{tabular}

\caption{Ablation study results of our proposed EAG-RL.}
\label{tab:ablation_results}
\end{table}

\section{Experiments}
In this section, we provide detailed information on the experimental setup, further analysis to validate the performance and rationality of \mname.

\subsection{Experimental Setup}
\noindent\textbf{Datasets.}
We conduct experiments on two real-world public EHR datasets: \textit{MIMIC-IV}~\cite{johnson2023mimic}, containing de-identified ICU records (2008–2019), and \textit{TJH}~\cite{yan2020interpretable}, comprising structured inpatient data with clinical annotations. Following prior work~\cite{gao2024comprehensive,zhu2024pyehr}, we apply temporal aggregation, LOCF imputation~\cite{wells2013strategies}, and sequence visits at the patient level. We select patients with at least two visits and use the last visit for prediction. Stratified train/validation/test splits are used. Full preprocessing details and dataset statistics are summarized in Appendix~E.

\noindent\textbf{Baselines.}
We compare \mname~with a wide range of baselines from three perspectives.
\begin{itemize}[leftmargin=*,label=\textbullet,noitemsep,topsep=2pt]
    \item \textbf{Prompt-based methods.} We include a \textit{Vanilla} baseline, where the model directly outputs the final answer without any intermediate reasoning. We also include \textit{Think then Answer}~\cite{guo2025deepseek}, which prompts the model to reason inside \texttt{<think>} tags before outputting a final answer in \texttt{<answer>}, and our proposed \textit{Question Decomposition} prompting, which encourages sub-question generation to support progressive reasoning.

    \item \textbf{Training-based methods.} We include standard SFT~\cite{ding20243ds,zhang2023huatuogpt,zelikman2022star} as a strong baseline for aligning models with task-specific data. To evaluate the benefit of \textit{Stage-1}, we compare against SFT directly. Additionally, to ensure a fair comparison in the RL stage, we compare \mname with GRPO\cite{shao2024deepseekmath} and DAPO~\cite{yu2025dapo} under the \textit{same Stage-1 initialization}, allowing us to isolate the effect of our reinforcement phase. Both GRPO and DAPO are state-of-the-art RL methods that have shown effectiveness in reasoning tasks.

    \item \textbf{Open-source LLMs and backbone variants.} To assess the generality of our method, we instantiate \mname{} on multiple backbone models, including Qwen2.5-7B-Instruct~\cite{yang2024qwen25}, LLaMA3.1-8B-Instruct~\cite{dubey2024llama}, and Qwen2.5-3B-Instruct~\cite{yang2025qwen3}. We also compare against several powerful open-source medical and reasoning LLMs such as HuatuoGPT-o1-7B~\cite{chen2024huatuogpt}, OpenBioLLM-8B~\cite{pal2024openbiollms}, and DeepSeek-R1-7B~\cite{guo2025deepseek}.
\end{itemize}

\noindent\textbf{Evaluation Metrics and Strategy.} We employ two widely used evaluation metrics to measure the performance, namely, Area Under the Receiver Operating Characteristic Curve (AUROC) and the Area Under the Precision-Recall Curve (AUPRC). Higher scores in these metrics indicate better predictive performance. Model selection is performed on the validation set. To assess variability, we apply bootstrapping with 100 resamples on the test set and report the mean and standard deviation in Table~\ref{tab:stage_results}.

\subsection{Experimental Results}
\noindent\textbf{Performance Comparison.}
Table~\ref{tab:stage_results} presents the performance of \mname and baselines across two datasets. Overall, across all evaluation metrics on the two datasets, especially AUPRC, which is the most informative primary evaluation metric when dealing with highly imbalanced datasets,  \textbf{\mname consistently outperforms the current state-of-the-art methods}. Compared to prompt-based approaches, \mname{} achieves an average improvement of 14.62\% across models and tasks, validating the intrinsic EHR reasoning enhancement from our two-stage framework. Notably, our \textit{Question Decomposition} also surpasses \textit{Think then Answer} baseline, suggesting that guiding reasoning via subquestion decomposition is a more effective prompting strategy in clinical contexts. Secondly, compared to vanilla SFT, the warm-up stage of \mname using expert-guided MCTS trajectories leads to consistent improvements. This underscores the importance of high-quality, stepwise initialization for aligning the LLM’s policy with clinical reasoning patterns. Finally, given the \textit{same Stage-1 initialization}, \mname consistently outperforms state-of-the-art RL methods. This fair comparison underscores the effectiveness of our reinforcement stage, which integrates expert attention and \textit{Entropy-Aware Adaptive Up Clipping} to guide exploration toward uncertain but informative clinical features.

\noindent\textbf{Ablation Study.} To investigate the effectiveness of each component in \mname, we construct several ablated variants. In \textit{w/o Stage-1}, we removes \textit{Expert-Guided Trajectory Distillation} and trains the model directly via RL. 
In \textit{w/o Stage-2}, we remove \textit{Attention-Aligned Policy Optimization}.
In \textit{w/o $\mathcal{R}_{\mathrm{att}}$}, we removes the \textit{Attention-Alignment Reward}.
In \textit{w/o $\epsilon(\tau)$},we disable the \textit{Entropy-aware Adaptive Up Clipping} during policy optimization. As shown in Table~\ref{tab:ablation_results}, removing either \textit{Stage-1} or \textit{Stage-2} leads to substantial performance degradation, confirming that both the expert-guided warm-up and the reinforcement refinement are essential for achieving strong EHR reasoning. In particular, the removal of\textit{ Stage-2} causes a larger drop, indicating that reward-based policy refinement plays a critical role beyond trajectory initialization. Furthermore, removing $\mathcal{R}_{\mathrm{att}}$ results in a noticeable decline in performance, validating the importance of leveraging expert attention as auxiliary supervision to guide the model’s focus toward clinically meaningful features. Similarly, without $\epsilon(\tau)$, the model fails to sufficiently explore uncertain yet informative patterns, leading to lower robustness and reduced predictive accuracy. Overall, the ablation results highlight the necessity of each component in \mname and demonstrate the synergistic benefits of expert-guided initialization and reward-aware policy optimization.

\begin{figure}[t]
  \centering
  \includegraphics[scale=0.105]{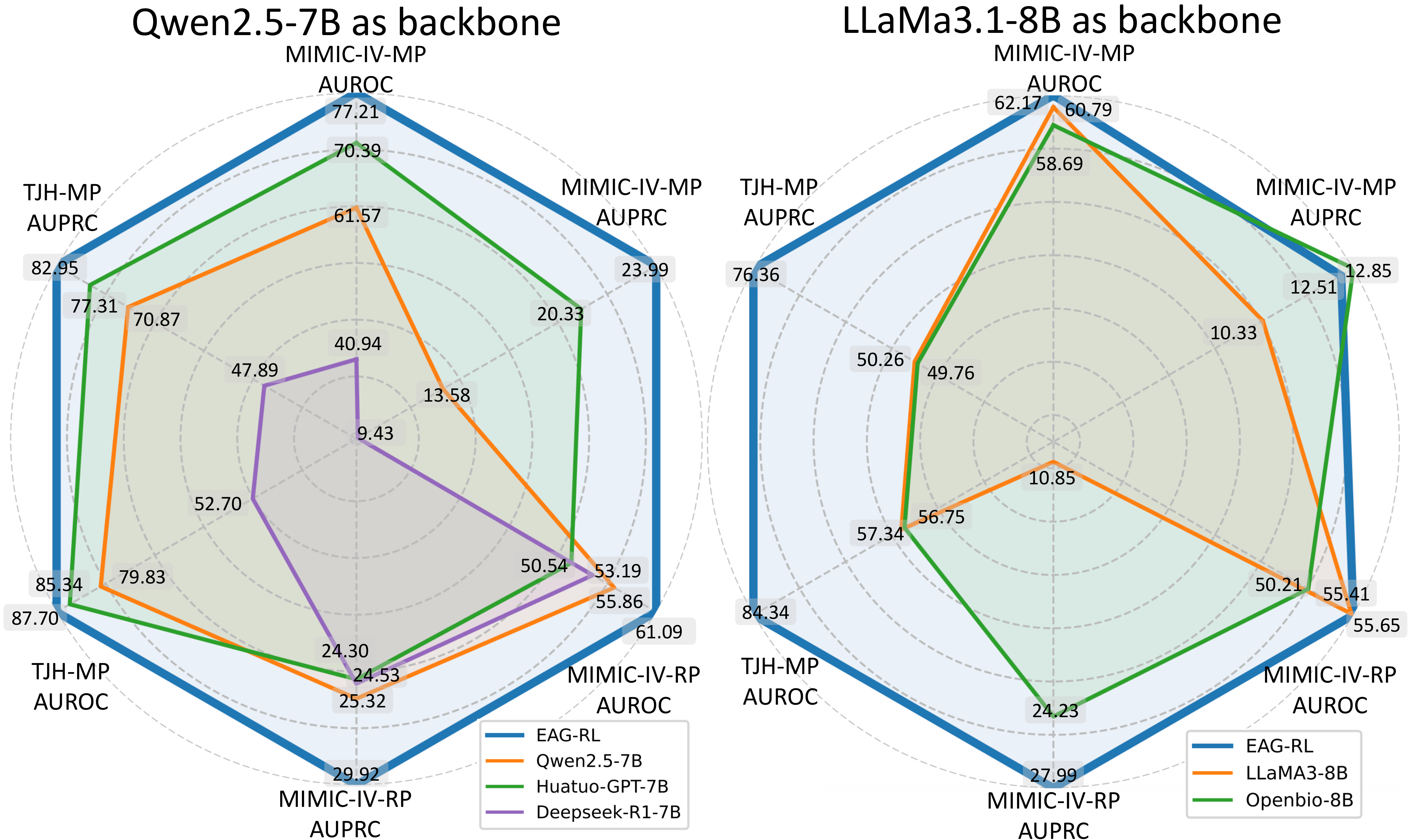}
  \caption{Performance comparison with open-source LLMs using Qwen2.5-7B (left) and LLaMA3.1-8B (right) as backbones. "MP" = mortality prediction; "RP" = readmission prediction.}
   \label{fig:data_scarcity}
\end{figure}

\begin{figure}[t]
  \centering
  \includegraphics[scale=0.35]{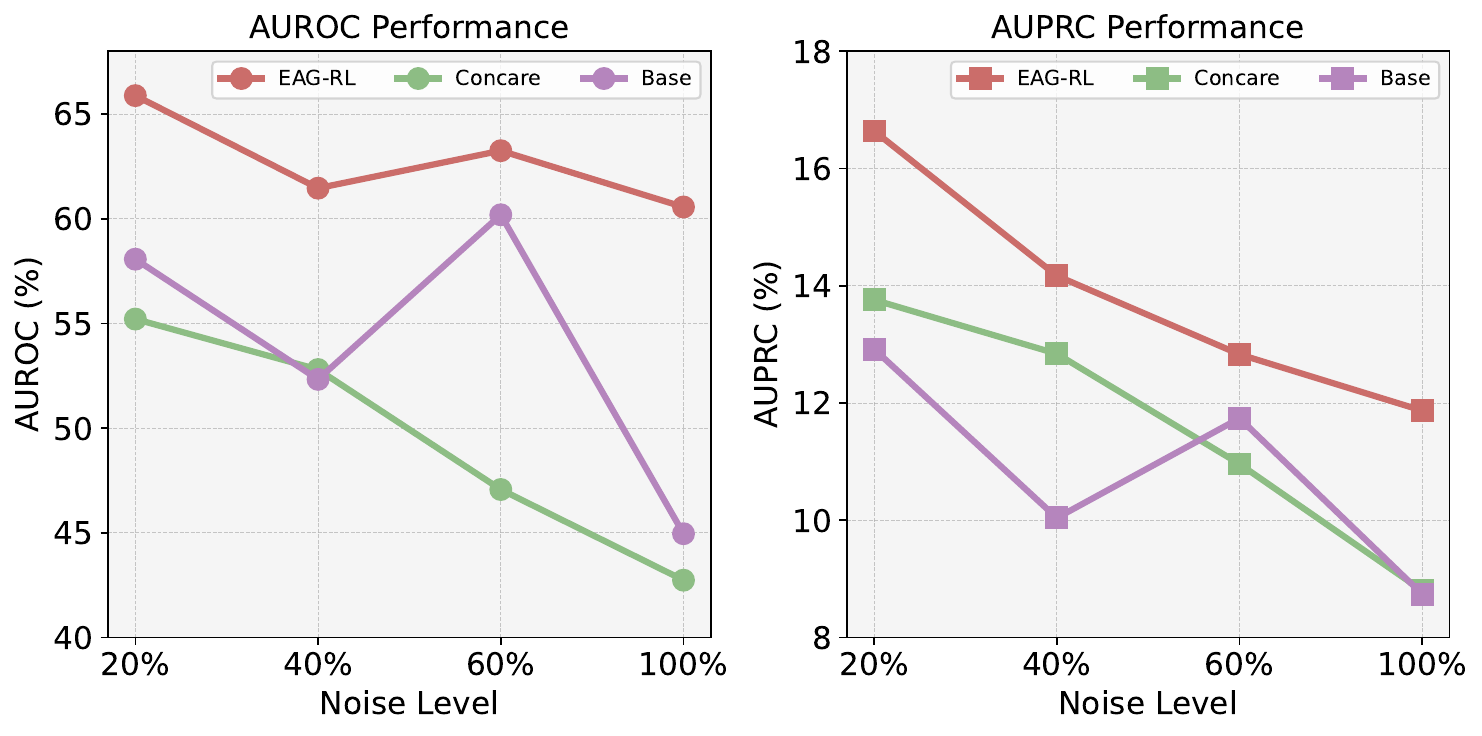}
  \caption{Robustness to feature order perturbation measured by AUROC (left) and AUPRC (right).}
   \label{fig:data_shuffle}
\end{figure}

\subsection{Analysis}
\noindent\textbf{EXP\#1: Intrinsic EHR Reasoning Capability.} To quantify how much our method enhances LLMs’ intrinsic EHR reasoning, we compare \mname{} with several competitive open-source models on mortality and readmission prediction tasks across TJH and MIMIC-IV datasets. Each task is evaluated using AUROC and AUPRC, resulting in six radar plot dimensions. Models are grouped by backbone for fair comparison. As shown in Figure~\ref{fig:data_scarcity}, \textbf{\mname{} consistently outperforms all competitive models across tasks and metrics.} Notably, \mname{} significantly surpasses HuatuoGPT-o1-7B and OpenBioLLM-8B, both trained on medical corpora or enhanced via domain-specific reinforcement learning, demonstrating the effectiveness of \mname in intrinsically strengthening EHR reasoning regardless of backbone.

\noindent\textbf{EXP\#2: Robustness Against Feature Order Perturbation.} To evaluate whether \mname{} captures clinically meaningful reasoning beyond superficial feature ordering, we conduct a robustness test on the \textit{mortality prediction} task by perturbing feature order at inference. This \textit{simulates real-world deployment scenarios} where structured EHR features may arrive in inconsistent orders due to variability in data collection, preprocessing, or integration pipelines. We compare against Concare, the expert EHR model, and Qwen2.5-7B, the untrained LLM backbone. For each MIMIC-IV test case, we randomly shuffle a proportion ($p\%$) of features while keeping the rest fixed. As shown in Figure~\ref{fig:data_shuffle}, Concare’s performance drops sharply under moderate perturbations, indicating its reliance on fixed input order. In contrast, \mname{} maintains robust performance across all disruption levels and continues to outperform the base model even under full permutation. These results suggest that \textbf{\mname{} learns semantically grounded, order-invariant reasoning strategies, enabling stronger generalization.} Such robustness is particularly valuable in clinical settings, where feature ordering often varies across institutions and systems.

\begin{figure}[t]
  \centering
  \includegraphics[scale=0.43]{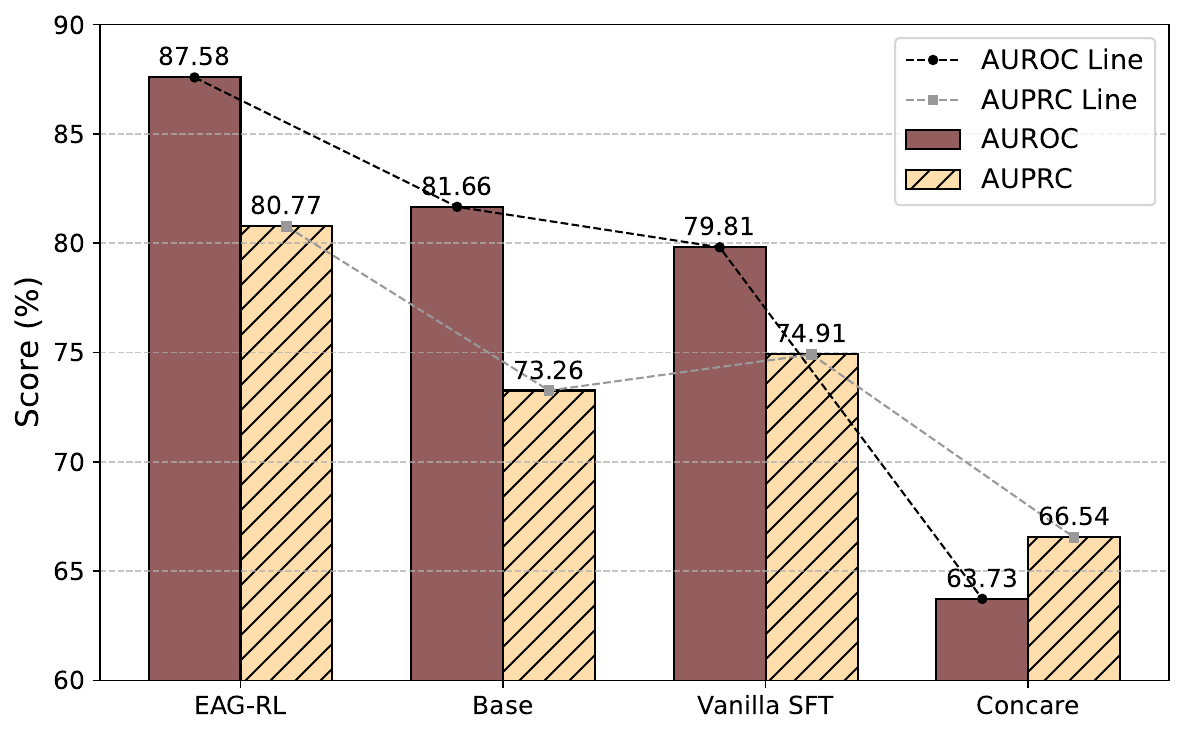}
  \caption{Cross-Dataset generalization from MIMIC-IV to TJH on mortality.}
   \label{fig:ood_experiment}
\end{figure}

\noindent\textbf{EXP\#3: Cross-Dataset Generalization.}
We assess \mname{}’s generalization via an out-of-distribution (OOD) test, training on MIMIC-IV and evaluating on TJH for \textit{mortality prediction}. TJH differs from MIMIC-IV in patient demographics and coding schemas, simulating deployment across heterogeneous medical environments. We compare against Concare, the untrained base model (Qwen2.5-7B), and Vanilla SFT. As shown in Figure~\ref{fig:ood_experiment}, \mname{} achieves the highest performance, significantly outperforming all baselines in both AUROC and AUPRC. These results suggest that \textbf{\mname{} captures transferable clinical patterns instead of relying on dataset-specific artifacts}.

\section{Conclusions and Future Works}

We present \textbf{EAG-RL}, a novel two stage training framework that enhances the intrinsic EHR reasoning ability of LLMs through expert-guided attention. Experiments show that \mname{} achieves an average improvement of 14.62\% across multiple EHR prediction tasks, while also improving robustness to input perturbations and generalization to unseen clinical domains. These results highlight the potential of \mname{} for real-world clinical deployment. In future work, we plan to explore richer forms of supervision beyond attention from expert EHR models, and extend our framework to incorporate multi-expert distillation to better capture diverse clinical reasoning patterns.

\pagestyle{empty}

\bibliography{aaai2026}

\newpage

\setcounter{secnumdepth}{0}
\section{Appendix}

\subsection{A. Related Work}
\subsubsection{Large Language Models and EHR}
Despite success in medical text processing~\cite{jahan2024comprehensive,chen2023large,zhou2023survey}, LLMs still underperform in EHR-based clinical prediction~\cite{brown2024not,chen2024clinicalbench,zhu2024clinicrealm}. Existing efforts~\cite{jiang2023graphcare,xu2025dearllm,xu2024ram,nguyen2024carer,zhu2024emerge} often adopt hybrid paradigms, where frozen LLMs provide medical priors while traditional deep learning(DL) models make final predictions~\cite{wang2023can}. Although effective, such approaches fail to intrinsically improve LLMs’ EHR reasoning and are limited by the generalization capacity of DL models. Prompt-based strategies~\cite{zhu2024clinicrealm} have also been explored, but rely on handcrafted heuristics and also fail to substantially improve the LLM’s intrinsic capabilities. Alternatively, supervised fine-tuning (SFT) has been used to adapt LLMs to EHR tasks~\cite{chen2024clinicalbench}, yet may overfit to narrow reasoning patterns~\cite{chu2025sft}, hindering generalization in complex scenarios.

\subsubsection{Large Language Models and RL}
Reinforcement Learning (RL)~\cite{kaelbling1996reinforcement,sutton1999reinforcement} has become a powerful paradigm for optimizing LLM behaviors through sequential decision-making. RL from human feedback (RLHF)~\cite{kaufmann2024survey} introduced reward-based policy tuning using human preference data. Subsequent methods such as DPO~\cite{rafailov2023direct} and SimPO~\cite{meng2024simpo} improve scalability by avoiding online sampling, though they often suffer from off-policy limitations. To further improve sample efficiency and exploration, Group Relative Policy Optimization (GRPO)~\cite{shao2024deepseekmath} eliminates the need for a critic by leveraging group-wise scoring, while Decoupled Clip and Dynamic sAmpling Policy Optimization (DAPO)~\cite{yu2025dapo} enhances GRPO by decoupling the clipping mechanism to better encourage policy exploration. Despite these advances, leveraging RL to intrinsically enhance LLMs’ reasoning over structured EHR data remains largely unexplored.

\subsection{B. Limitations}
While our proposed EAG-RL framework demonstrates promising results on EHR reasoning tasks, several limitations remain. First, the scope of our evaluation is limited to two EHR datasets, and scaling up to multiple EHR datasets with two-stage training, incorporating a more diverse set of EHR reasoning data, would help construct a more foundational EHR model. Second, due to computational resource constraints, we evaluate the performance using models with up to 8B parameters, and scaling to larger models could offer additional insights into the efficacy of EAG-RL. Finally, we currently consider the knowledge from a single expert EHR model; exploring richer forms of supervision beyond expert attention and incorporating multi-expert distillation could better capture diverse clinical reasoning patterns.

\subsection{C. Algorithm Details}
\label{sec: algorithm} 
In this section, we provide a detailed introduction to the algorithms used in our framework.

\subsubsection{Expert-Guided Trajectory Distillation}
We present the full pseudocode of our \textit{Expert-Guided Trajectory Distillation} process in Algorithm~\ref{alg:expert_mcts}.
\begin{algorithm*}[t]
\centering
\caption{Expert-Guided Trajectory Distillation}
\label{alg:expert_mcts}
\begin{minipage}{1\linewidth}
\normalsize
\begin{algorithmic}[1]
\Require Patient EHR data $\mathbf{X}$, Initial LLM $\mathcal{M}$, Expert EHR model $\mathcal{M}_{\text{exp}}$, Max depth $T$, Rollout number $N$, Exploration weight $\lambda_{1}$, Attention align weight $\lambda_{2}$, Number of candidate subquestions $d$
\State \textcolor{gray}{\# 1: Expert-Guided MCTS for Trajectory Construction}
\State Initialize: $Q(s, a)$, $N(s)$, $A(s)$, $c(s, a)$, $r(s, a)$
\State Generate initial prompt $\mathcal{P}_{\text{QD}}(\mathbf{X})$, set $s_0 \gets \mathcal{P}_{\text{QD}}(\mathbf{X})$
\State Initialize trajectory set $\mathcal{T} \gets \emptyset$
\For{$n \gets 1$ to $N$}
    \State $t \gets 0$, $s_t \gets s_0$
    \While{$N(s_t) > 0$} \Comment{Selection}
        \State $N(s_t) \gets N(s_t) + 1$
        \State $a_t \gets \arg\max_{a \in A(s_t)} \left[ Q(s_t, a) + \lambda_{1} \sqrt{\frac{\ln N(s_t)}{N(c(s_t, a))}} \right]$
        \State $s_{t+1} \gets c(s_t, a_t)$, $t \gets t + 1$
    \EndWhile
    \While{$s_t$ is not terminal \textbf{and} $t \leq T$} \Comment{Expansion \& Simulation}
        \For{$i \gets 1, \dots, d$}
            \State Sample $q_t^{(i)} \sim \mathcal{M}(\cdot \mid s_t)$
            \State Sample $a_q^{(i)} \sim \mathcal{M}(q_t^{(i)} \mid s_t \cup q_t^{(i)})$
            \State Let $a_t^{(i)} \gets (q_t^{(i)}, a_q^{(i)})$
            \State $s_{t+1}^{(i)} \gets \text{expand}(s_t, a_t^{(i)})$
            \State $r_t^{(i)} \gets \mathcal{M}_{\mathcal{P}_{\text{H}}}(\tau_{1:t} \cup \{(q_t^{(i)}, \varnothing)\})$
            \State Update:
            \State \quad $A(s_t) \gets A(s_t) \cup \{a_t^{(i)}\}$
            \State \quad $c(s_t, a_t^{(i)}) \gets s_{t+1}^{(i)}$
            \State \quad $r(s_t, a_t^{(i)}) \gets r_t^{(i)}$

        \EndFor
        \State $a_{t+1} \gets \arg\max_{a \in A(s_t)} r(s_t, a)$
        \State $s_{t+1} \gets c(s_t, a_{t+1})$, $t \gets t + 1$
    \EndWhile
    \If{$s_t$ is terminal} \Comment{Backpropagation}
        \State Extract prediction $(\hat{y}, \mathcal{C})$ from final answer
        \State $\mathcal{R}_{\mathrm{cls}} \gets \log(y^\star \cdot \hat{y} + (1 - y^\star)(1 - \hat{y})) + \Delta$
        \State $\mathcal{R}_{\mathrm{att}} \gets \frac{|\mathcal{C} \cap \mathcal{C}_{\mathrm{exp}}|}{|\mathcal{C} \cup \mathcal{C}_{\mathrm{exp}}|}$
        \State $\mathcal{R} \gets \lambda_{2} \cdot \mathcal{R}_{\mathrm{cls}} + (1 - \lambda_{2}) \cdot \mathcal{R}_{\mathrm{att}}$
        \For{$t' \gets t$ down to $1$}
            \State Update $Q(s_{t'}, a_{t'})$ with $\mathcal{R}$
            \State $N(s_{t'}) \gets N(s_{t'}) + 1$
        \EndFor
    \EndIf
    \State Save trajectory $\tau$ to $\mathcal{T}$
\EndFor
\vspace{1mm}
\State \textcolor{gray}{\# 2: Trajectory-Level Supervised Fine-Tuning}
\State Construct $\mathcal{D}_{\text{SFT}} \gets \{\tau_i\}_{i=1}^k$ from top-$k$ reward trajectories
\State Update $\mathcal{M}$ by minimizing SFT loss:
\State \quad $\mathcal{L}_{\text{SFT}} = - \sum_{\tau \in \mathcal{D}_{\text{SFT}}} \log \mathcal{M}(\tau \mid \mathcal{P}_{\text{QD}}(\mathbf{X}))$

\end{algorithmic}
\end{minipage}
\end{algorithm*}

\subsubsection{Attention-Aligned Policy Optimization}
\label{appendix:grpo_details}

We adopt Group Relative Policy Optimization (GRPO)~\cite{shao2024deepseekmath} as the backbone RL algorithm. GRPO stabilizes policy updates through group-normalized rewards and a clipped surrogate loss. In this section, we detail the loss formulation and our integration of entropy-adaptive clipping.

\paragraph{Group Sampling.}
For each input query \( q \), we sample a group of \( G \) trajectories from the current policy:
\begin{equation}
\{ \tau_i \}_{i=1}^G \sim \pi_{\theta_{\text{old}}}(\cdot \mid q).
\label{eq:group_sampling}
\end{equation}

\paragraph{Group-Normalized Advantage.}
Given the reward \( R_i \) for each trajectory, we compute the group-relative advantage:
\begin{equation}
\begin{aligned}
\hat{A}_i &= \frac{R_i - \mu_R}{\sigma_R}, \\
\mu_R &= \frac{1}{G} \sum_{j=1}^G R_j, \\
\sigma_R &= \sqrt{ \frac{1}{G} \sum_{j=1}^G (R_j - \mu_R)^2 }.
\end{aligned}
\label{eq:group_advantage}
\end{equation}

\paragraph{Importance Ratio.}
For each token \( o_{i,t} \) in trajectory \( \tau_i \), the importance sampling ratio is:
\begin{equation}
r_{i,t} = \frac{\pi_\theta(o_{i,t} \mid q, o_{i,<t})}{\pi_{\theta_{\text{old}}}(o_{i,t} \mid q, o_{i,<t})}.
\label{eq:importance_ratio}
\end{equation}

\paragraph{Asymmetric Clipping Function.}
Following DAPO~\cite{yu2025dapo}, we adopt an asymmetric clipping function defined as:
\begin{equation}
\phi(r; \varepsilon, \epsilon(\tau)) = \max(1 - \varepsilon,\ \min(r,\ 1 + \epsilon(\tau))),
\label{eq:asymmetric_clip}
\end{equation}
where \( \varepsilon \) is the fixed lower bound and \( \epsilon(\tau) \) is the adaptive upper bound derived from token-level entropy.

\paragraph{Token-Level Surrogate Loss.}
We define the clipped objective at each token \( t \) in trajectory \( \tau_i \) as:
\begin{equation}
\mathcal{L}_{i,t} = \min \left( r_{i,t} \cdot \hat{A}_i,\ \phi(r_{i,t}; \varepsilon, \epsilon(\tau_i)) \cdot \hat{A}_i \right).
\label{eq:token_loss}
\end{equation}

\paragraph{GRPO Objective.}
The overall GRPO-style objective becomes:
\begin{equation}
\mathcal{J}(\theta) = \mathbb{E}_{q \sim \mathcal{D}} \left[ \frac{1}{G} \sum_{i=1}^G \frac{1}{|\tau_i|} \sum_{t=1}^{|\tau_i|} \mathcal{L}_{i,t} \right],
\label{eq:grpo_loss_full}
\end{equation}
where \( \mathcal{D} \) denotes the training dataset.

\clearpage
\subsection{D. Prompt Details}
\label{sec: prompts} 
In this section, we provide a detailed introduction to the prompts used in our framework. For each task (e.g., mortality prediction, readmission prediction), we present both the \textit{Question Decomposition Prompts} ~($\mathcal{P}_{\text{QD}}$), which guides the model to generate subquestions and intermediate reasoning steps, and the \textit{valuation prompt}~($\mathcal{P}_{\text{H}}$), which is used to assess the clinical utility of candidate subquestions.

The \textit{Prompt for Mortality Prediction} and \textit{Prompt for Readmission Prediction} correspond to task-specific \textit{Question Decomposition Prompts}~($\mathcal{P}_{\text{QD}}$) for the mortality and readmission tasks, respectively.

\begin{tcolorbox}
[colback=lightgray!20,
colframe=darkgray!80,
fontupper=\small,
title=Prompt for Mortality Prediction]
\label{tab:prompt_mortality}
You are tasked with predicting in-hospital mortality based on patient EHR data. Please strictly follow the reasoning format below without omission:\\

\textbf{\textit{Reasoning Format:}}

\texttt{<Subquestion 1>} ... \texttt{</Subquestion 1>} \\
\texttt{<Answer 1>} ... \texttt{</Answer 1>} \\

\texttt{<Subquestion 2>} ... \texttt{</Subquestion 2>} \\
\texttt{<Answer 2>} ... \texttt{</Answer 2>} \\

... \\
\texttt{<Final subquestion>} Now we can determine the likelihood of the patient not surviving their hospital stay. \texttt{</Final subquestion>} \\
\texttt{<Important Features>} feature1, feature2, feature3, feature4, feature5 \texttt{</Important Features>} \\
\texttt{<Final answer>} 0.XX \texttt{</Final answer>}\\

\textbf{\textit{Scoring Guidelines:}}

\texttt{Score 1} = high death likelihood; \texttt{0} = high survival likelihood. \\
Use \texttt{0.5} only for genuinely conflicting or insufficient data. \\

\textbf{\textit{Example Output:}}

\texttt{<Subquestion 1>} Does the patient show signs of infection? \texttt{</Subquestion 1>} \\
\texttt{<Answer 1>} Yes, elevated WBC count and persistent fever. \texttt{</Answer 1>} \\

\texttt{<Subquestion 2>} Are there signs of hypotension or sepsis? \texttt{</Subquestion 2>} \\
\texttt{<Answer 2>} Blood pressure is critically low, SOFA score is elevated. \texttt{</Answer 2>} \\

\texttt{<Final subquestion>} Now we can determine the likelihood of the patient not surviving their hospital stay. \texttt{</Final subquestion>} \\

\texttt{<Important Features>} age 85, septic shock, SOFA=11, low MAP, abnormal labs \texttt{</Important Features>} \\

\texttt{<Final answer>} 0.91 \texttt{</Final answer>}
\end{tcolorbox}

\begin{tcolorbox}
[colback=lightgray!20,colframe=darkgray!80,title=Prompt for Readmission Prediction]
\label{tab:prompt_readmission}
You are tasked with predicting whether the patient will be readmitted within two weeks based on EHR data. Please strictly follow the reasoning format below without omission:\\

\textbf{\textit{Reasoning Format:}}

\texttt{<Subquestion 1>} ... \texttt{</Subquestion 1>} \\
\texttt{<Answer 1>} ... \texttt{</Answer 1>} \\

\texttt{<Subquestion 2>} ... \texttt{</Subquestion 2>} \\
\texttt{<Answer 2>} ... \texttt{</Answer 2>} \\

... \\
\texttt{<Final subquestion>} Now we can determine the likelihood of the patient being readmitted to the hospital within 30 days of discharge. \texttt{</Final subquestion>} \\
\texttt{<Important Features>} feature1, feature2, feature3, feature4, feature5 \texttt{</Important Features>} \\
\texttt{<Final answer>} 0.XX \texttt{</Final answer>}\\

\textbf{\textit{Scoring Guidelines:}}

\texttt{Score 1} = high readmission likelihood; \texttt{0} = low readmission likelihood. \\
Use \texttt{0.5} only for genuinely conflicting or insufficient data. \\

\textbf{\textit{Example Output:}}

\texttt{<Subquestion 1>} Has the patient experienced frequent ER visits in the past month? \texttt{</Subquestion 1>} \\
\texttt{<Answer 1>} Yes, three ER visits due to heart failure exacerbations. \texttt{</Answer 1>} \\

\texttt{<Subquestion 2>} Does the discharge summary mention unresolved conditions or poor follow-up plans? \texttt{</Subquestion 2>} \\
\texttt{<Answer 2>} The discharge note highlights inadequate outpatient follow-up and ongoing respiratory issues. \texttt{</Answer 2>} \\

\texttt{<Final subquestion>} Now we can determine the likelihood of the patient being readmitted to the hospital within two weeks. \texttt{</Final subquestion>} \\

\texttt{<Important Features>} recent ER visits, CHF, no follow-up plan, ongoing symptoms, poor discharge coordination \texttt{</Important Features>} \\

\texttt{<Final answer>} 0.87 \texttt{</Final answer>}
\end{tcolorbox}

Similarly, the \textit{Prompt for Sub-question Usefulness Evaluation} corresponds to the valuation prompt~($\mathcal{P}_{\text{H}}$), which is used to assess the utility of each generated sub-question in the reasoning trajectory.

\begin{tcolorbox}
[colback=lightgray!20,colframe=darkgray!80,title=Prompt for Sub-question Usefulness Evaluation]
\label{tab:prompt_subq_score}
Given a case and its generated subquestions and answers, evaluate the usefulness of the \textbf{last sub-question} (i.e., the one not yet answered) for answering the original clinical prediction question. \\

Provide a score from 1 to 100, where: \\
\texttt{1} = completely useless; \\
\texttt{100} = extremely useful. \\

Also, provide a brief reason for your score. \\

\textbf{\textit{Criteria for a good sub-question:}} \\
- Introduces new, clinically relevant information not already covered \\
- Focuses on abnormal values, trends, or critical physiological signals \\
- Helps move closer to answering the main question (e.g., predicting mortality) \\
- Encourages analysis of multiple features together (e.g., low pH + high RR) \\
- Helps uncover potential risks that may not be obvious from single variables \\

\textbf{\textit{Criteria for a poor sub-question:}} \\
- Repeats prior questions or rephrases the main question \\
- Focuses on static or irrelevant features (e.g., height) \\
- Ignores previously discussed risk factors \\
- Adds no meaningful reasoning value or clinical insight \\

\textbf{\textit{Output Format:}} \\
\texttt{The usefulness score is: <score>} \\
\texttt{<brief explanation>}
\end{tcolorbox}


\subsection{E. EHR Dataset}
\label{EHR Dataset}
In this section, we provide a detailed introduction of the datasets used in our study, along with the preprocessing procedures applied to the structured EHR data.
\subsubsection{Dataset Description}
Our experiments are conducted on two real-world EHR datasets: the \textbf{TJH dataset}~\cite{yan2020interpretable} and the \textbf{MIMIC-IV dataset}~\cite{johnson2023mimic}. The \textit{TJH dataset} comprises structured EHR data collected from Tongji Hospital, affiliated with Tongji Medical College. It includes a wide range of clinical variables and has been publicly released by the HAIR Lab for research on patient outcome prediction. The \textit{MIMIC-IV dataset}, developed by the MIT Laboratory for Computational Physiology, contains de-identified health records from the Beth Israel Deaconess Medical Center. In this work, we use version 3.1 of its structured EHR component, which includes demographics, vital signs, laboratory tests, medication prescriptions, and more. This dataset enables comprehensive modeling of patient trajectories across various clinical tasks.

\subsubsection{Data Preprocessing}

We preprocess both datasets using the open-source pipeline, following standardized practices established in prior benchmarks~\cite{gao2024comprehensive,zhu2024pyehr}.  For \textit{MIMIC-IV dataset}, time-stamped events within the same ICU stay are aggregated into daily records. For stays longer than seven days, only the most recent seven days are retained, with earlier records summarized. Missing values are imputed using the Last Observation Carried Forward (LOCF) method~\cite{wells2013strategies} to preserve temporal continuity. The \textit{TJH dataset} undergoes the same temporal alignment and imputation strategy. All records are then chronologically ordered and structured into visit-level sequences for downstream reasoning tasks. To ensure consistent evaluation, we partition each dataset into training, validation, and test sets via stratified sampling, with a fixed test set used across all experiments. Summary statistics are presented in Table~\ref{tab:dataset_stats}.

\begin{table*}[!ht]
\small
\centering
\caption{\textit{Statistics of the TJH dataset and MIMIC-IV dataset.} ``Re.'' stands for Readmission, indicating patients who are readmitted to the ICU within 30 days of discharge, while ``No Re.'' represents patients who are not readmitted.}
\label{tab:dataset_stats}
\begin{tabular}{lccccccccc}
\toprule
\multicolumn{1}{c}{\textbf{Dataset}} & \multicolumn{3}{c}{\textbf{TJH}} & \multicolumn{5}{c}{\textbf{MIMIC-IV EHR}} \\
\cmidrule(lr){2-4} \cmidrule(lr){5-9}
                & Total & Alive & Dead & Total & Alive & Dead & Re. & No Re. \\
\midrule
\multicolumn{9}{c}{\textit{Test Set Statistics}} \\
\midrule
\# Patients     & 200 & 109 & 91 & 200 & 183 & 17 & 53 & 147 \\
\# Total visits & 967 & 601 & 366 & 801 & 717 & 84 & 274 & 527 \\
\# Avg. visits  & 4.8 & 5.5 & 4.0 & 4.0 & 3.9 & 4.9 & 5.2 & 3.6 \\
\midrule
\multicolumn{9}{c}{\textit{Training Set Statistics}} \\
\midrule
\# Patients     & 140 & 75 & 65 & 8750 & 8028 & 722 & 2112 & 6638 \\
\# Total visits & 641 & 395 & 246 & 33423 & 30117 & 3306 & 10448 & 22975 \\
\# Avg. visits  & 4.6 & 5.3 & 3.8 & 3.8 & 3.8 & 4.6 & 4.9 & 3.5 \\
\midrule
\multicolumn{9}{c}{\textit{Validation Set Statistics}} \\
\midrule
\# Patients     & 21 & 11 & 10 & 1250 & 1147 & 103 & 305 & 945 \\
\# Total visits & 96 & 54 & 42 & 4685 & 4176 & 509 & 1522 & 3163 \\
\# Avg. visits  & 4.6 & 4.9 & 4.2 & 3.7 & 3.6 & 4.9 & 5.0 & 3.3 \\
\bottomrule
\end{tabular}
\end{table*}

\subsection{F. Detailed Experimental Setup}
\label{Detailed Experimental Setup}

\subsubsection{Compared Methods.}


We comprehensively evaluate our approach against several state-of-the-art methods spanning prompt-based methods, training-based methods, specialized open-source LLMs, and expert small-scale models. 

\textbf{Prompt-Based Methods.} \textit{Vanilla} directly prompts the model to output the final answer without intermediate reasoning, serving as a basic baseline for comparison. \textit{Think-then-Answer}~\cite{guo2025deepseek} explicitly distinguishes between reasoning and final predictions by encapsulating reasoning steps within dedicated \texttt{<think>...</think>} tags and final answers within \texttt{<answer>...</answer>} tags. \textit{Question Decomposition}, inspired by clinical reasoning methodologies, strategically decomposes complex EHR prediction tasks into sequential sub-questions, simplifying the inference process into manageable components.

\textbf{Training-Based Methods.} \textit{Supervised Fine-Tuning (SFT)} serves as a robust baseline, employing balanced datasets of 1,000 samples (500 positive, 500 negative) extracted from patient EHR data, training models for direct predictions without explicit intermediate reasoning. To address the limited size of the TJH dataset, we employ dataset replay to enhance training stability. \textit{Group Relative Policy Optimization (GRPO)}~\cite{shao2024deepseekmath} stabilizes policy updates by normalizing rewards across sampled trajectory groups, promoting robust learning. For this method, we use a \textit{Classification reward} \( \mathcal{R}_{\mathrm{cls}} \in \mathbb{R} \) to guide optimization. \textit{Dynamic sAmpling Policy Optimization (DAPO)}~\cite{yu2025dapo} employs distinct clipping mechanisms and dynamic sampling to mitigate entropy collapse and encourage exploration of diverse trajectories. Similar to GRPO, we use a \textit{Classification reward} \( \mathcal{R}_{\mathrm{cls}} \in \mathbb{R} \) in DAPO to direct the optimization process. Additionally, to ensure a fair comparison in the RL stage, we compare \mname with GRPO\cite{shao2024deepseekmath} and DAPO~\cite{yu2025dapo} under the \textit{same Stage-1 initialization}, allowing us to isolate the effect of our reinforcement phase. 

\textbf{Open-Source LLMs and Variants.} \textit{HuatuoGPT-o1-7B}~\cite{chen2024huatuogpt} is trained using RL specifically on healthcare domain data, optimizing its medical reasoning capabilities. \textit{OpenBioLLM-8B}~\cite{pal2024openbiollms} is pre-trained on biomedical texts to deeply understand biological and clinical contexts. \textit{DeepSeek-R1-7B}~\cite{guo2025deepseek} is fine-tuned on high-quality reasoning SFT data distilled from DeepSeek-R1~\cite{guo2025deepseek}, which enhances its general reasoning abilities across various tasks. It has demonstrated strong performance on a wide range of reasoning tasks.

\textbf{Expert EHR Model.} \textit{Concare}~\cite{ma2020concare}, a well-established EHR-specific expert model based on the Transformer architecture. It serves as our expert EHR model for \textbf{EAG-RL}, providing essential attention-based guidance for clinical reasoning.

Collectively, these methods encompass a comprehensive spectrum of contemporary optimization and modeling approaches, facilitating rigorous comparative analysis and clearly highlighting the unique advantages of our proposed EAG-RL framework.

\subsubsection{Evaluation Metrics.}

We employ standard evaluation metrics for each prediction task. For the classification tasks, including in-hospital mortality and 30-day readmission prediction, we report the Area Under the ROC Curve (AUROC) and the Area Under the Precision-Recall Curve (AUPRC), both widely used to assess performance on imbalanced binary classification problems. Model selection is performed on the validation set. 

\subsubsection{Experimental Implementation.}

The hyperparameters for our method are configured as follows. In the \textit{Expert-Guided Trajectory Distillation} stage, the rollout number is set to $N=8$, and the maximum reasoning depth is $T=7$. The exploration weight in the UCT formula is $\lambda_{1}=1.0$. The terminal reward combines classification reward ($\mathcal{R}_{\mathrm{cls}}$) and attention alignment reward ($\mathcal{R}_{\mathrm{att}}$) using a weighting parameter $\lambda_{2}=0.6$. At each expansion step, we generate $d=3$ candidate sub-questions. To construct the dataset $\mathcal{D}_{\text{SFT}}$ for Stage-1 fine-tuning, we select the top $k=2$ trajectories with the highest rewards. 
In the \textit{Attention-Aligned Policy Optimization} stage, we set the LoRA rank to 64 and the scaling factor to 32, respectively. The rollout number during training is 8. We optimize our model parameters using the Adam optimizer with a learning rate of $3\times10^{-6}$. To balance exploration and exploitation during policy updates, we set the GRPO clipping lower bound as $\varepsilon=0.2$, and adaptively set the upper bound within the range $\epsilon_{\min}=0.2$ to $\epsilon_{\max}=0.4$ based on the \textit{Entropy-Aware Adaptive Up Clipping}.

The experiments are conducted on a Linux operating system running Ubuntu 22.04.5 LTS, with kernel version 6.8.0 on an x86\_64 architecture. The computational resources consist of 4 NVIDIA 4090 GPUs, each with 48GB of memory. The implementation uses Python 3.10.0, PyTorch 2.6.0+cu124, Transformers 4.51.1, and VERL 0.4.0~\cite{sheng2025hybridflow}. All experiments use a fixed random seed (42) for dataset splitting to ensure reproducibility. Each experiment is run once, and results are reported with variance estimates computed via bootstrap resampling with 100 samples.

\subsection{G. Case Study}

To qualitatively demonstrate how EAG-RL improves the LLM's reasoning process, we present a case study on a patient from the test set. We analyze a specific instance where the baseline LLM makes an incorrect prediction. In contrast, the model trained with our EAG-RL framework successfully corrects the prediction.
The core observation is that EAG-RL effectively steers the LLM's attention toward clinically salient features. This is evidenced by a significant increase in the overlap between the features identified as important by our model and those highlighted by the expert model. This shift in attention directly contributes to the improved predictive accuracy, as detailed in Table~\ref{table:case_study}.

\subsection{H. Code Availability}

The code for this research has been submitted alongside this appendix, with detailed usage instructions included. It will be made publicly available to support further research and facilitate learning within the community.

\subsection{I. Ethics Statement}
This work adheres to high ethical standards for AI research in healthcare. All experiments are conducted on publicly available, de-identified EHR datasets (MIMIC-IV and TJH), ensuring full compliance with data privacy regulations and preventing exposure of any protected health information (PHI).
Our proposed method, EAG-RL, is intended solely as a research framework to enhance the reasoning ability of LLMs over structured EHR data. It is not designed to replace human clinical judgment, provide medical advice, or serve as a diagnostic system in real-world clinical settings. All generated outputs should be interpreted and validated by qualified medical professionals.
We recognize the importance of transparency and reproducibility in high-stakes domains such as healthcare. To facilitate future research and independent verification, we commit to releasing our source code and data processing scripts upon publication.
To promote fairness and generalizability, our method is evaluated across diverse EHR benchmarks, and we acknowledge the potential limitations of model performance across demographic or institutional shifts. Future work may explore additional safeguards and fairness auditing mechanisms to ensure equitable clinical applicability.

\renewcommand{\arraystretch}{2.4}
\onecolumn
\begin{center}
\begin{small}
\begin{longtable}{
  >{\centering\arraybackslash}m{2cm}
  >{\raggedright\arraybackslash}m{11.25cm}
}

\caption{\label{table:case_study} Case study on EHR mortality prediction.}\\
\hline
\textbf{Prompt} & 
\begin{minipage}{11.25cm} %
\vspace{5pt}
I will provide you with longitudinal medical information for a patient. The data covers 2 visits that occurred at 2119-12-24, 2119-12-24.
Each clinical feature is presented as a list of values, corresponding to these visits. Missing values are represented as `NaN` for numerical values and "unknown" for categorical values. \newline

\textbf{\textit{Patient Background:}}

- Sex: female\newline
- Age: 81 years \newline

You are tasked with predicting in-hospital mortality based on patient EHR data. Please strictly follow the reasoning format below without omission:\\

\textbf{\textit{Reasoning Format:}}

\texttt{<Subquestion 1>} ... \texttt{</Subquestion 1>} \\
\texttt{<Answer 1>} ... \texttt{</Answer 1>} \\
\texttt{<Subquestion 2>} ... \texttt{</Subquestion 2>} \\
\texttt{<Answer 2>} ... \texttt{</Answer 2>} \\
... \\
\texttt{<Final subquestion>} Now we can determine the likelihood of the patient not surviving their hospital stay. \texttt{</Final subquestion>} \\
\texttt{<Important Features>} feature1, feature2, feature3, feature4, feature5 \texttt{</Important Features>} \\
\texttt{<Final answer>} 0.XX \texttt{</Final answer>}\\

\textbf{\textit{Scoring Guidelines:}}

\texttt{Score 1} = high death likelihood; \texttt{0} = high survival likelihood. \\
Use \texttt{0.5} only for genuinely conflicting or insufficient data. \\

\textbf{\textit{Example Output:}}

\texttt{<Subquestion 1>} Does the patient show signs of infection? \texttt{</Subquestion 1>} \\
\texttt{<Answer 1>} Yes, elevated WBC count and persistent fever. \texttt{</Answer 1>} \\
\texttt{<Subquestion 2>} Are there signs of hypotension or sepsis? \texttt{</Subquestion 2>} \\
\texttt{<Answer 2>} Blood pressure is critically low, SOFA score is elevated. \texttt{</Answer 2>} \\
\texttt{<Final subquestion>} Now we can determine the likelihood of the patient not surviving their hospital stay. \texttt{</Final subquestion>} \\
\texttt{<Important Features>} age 85, septic shock, SOFA=11, low MAP, abnormal labs \texttt{</Important Features>} \\
\texttt{<Final answer>} 0.91 \texttt{</Final answer>} \\

Now, please analyze and predict for the following patient:\\

\textbf{\textit{Clinical Features Over Time:}}

- Capillary refill rate: [0.0, 1.0]\\
- Glascow coma scale eye opening: [Spontaneously, Spontaneously]\\
- Glascow coma scale motor response: [Obeys Commands, Obeys Commands]\\
- Glascow coma scale total: [15.0, 15.0]\\
- Glascow coma scale verbal response: [Oriented, Oriented]\\
- Diastolic blood pressure: [57.0, 70.0]\\
- Fraction inspired oxygen: [100.0, 50.0]\\
- Glucose: [484.0, 178.0]\\
- Heart Rate: [88.0, 80.0]\\
- Height: [157.0, NaN]\\
- Mean blood pressure: [82.0, 90.0]\\
- Oxygen saturation: [99.0, 99.0]\\
- Respiratory rate: [25.0, 24.0]\\
- Systolic blood pressure: [96.0, 94.0]\\
- Temperature: [37.6, 37.3]\\
- Weight: [62.2, 119.8]\\
- pH: [7.13, 7.16]\\
\end{minipage}
\\ \hline
\textbf{Ground Truth} & 1 (Mortality) \\ \hline
\textbf{Expert Features} & 'Age \& Sex', 'pH', 'Glucose', 'Height', 'Systolic blood pressure' \\ \hline
\textbf{Baseline LLM Output} & 
\textbf{Prediction}: 0.25 → \textbf{\color{red}{Incorrect}} \quad | \quad \textbf{Feature Overlap with Expert}: \textbf{2/5}
\newline
\newline
\texttt{<Subquestion 1>}Is the patient's capillary refill rate normal?\texttt{</Subquestion 1>}
\newline
\texttt{<Answer 1>}The capillary refill rate is normal, indicating adequate perfusion.\texttt{</Answer 1>}
\newline
\newline
\texttt{<Subquestion 2>}How does the patient's Glasgow Coma Scale (GCS) score change over time?\texttt{</Subquestion 2>}
\newline
\texttt{<Answer 2>}The GCS score remains stable at 15, indicating full consciousness.\texttt{</Answer 2>}
\newline
\newline
\texttt{<Subquestion 3>}What is the trend in diastolic blood pressure?\texttt{</Subquestion 1>}
\newline
\texttt{<Answer 3>}Diastolic blood pressure improves from critically low to within normal range.\texttt{</Answer 3>}
\newline
\newline
\texttt{<Subquestion 4>}How does the fraction of inspired oxygen (FiO2) change over time?\texttt{</Subquestion 1>}
\newline
\texttt{<Answer 4>}FiO2 decreases from 100\% to 50\%, suggesting improvement in oxygenation.\texttt{</Answer 4>}
\newline
\newline
\texttt{<Subquestion 5>}What is the trend in glucose levels?\texttt{</Subquestion 1>}
\newline
\texttt{<Answer 5>}Glucose levels significantly decrease from very high to moderate, indicating possible intervention.\texttt{</Answer 5>}
\newline
\newline
\texttt{<Subquestion 6>}How does the patient's pH change over time?\texttt{</Subquestion 6>}
\newline
\texttt{<Answer 6>}pH improves slightly but remains below normal, indicating metabolic acidosis.\texttt{</Answer 6>}
\newline
\newline
\texttt{<Final subquestion>}Now we can determine the likelihood of the patient not surviving their hospital stay.\texttt{</Final subquestion>}
\begin{tcolorbox}[colback=red!10!white,colframe=white,boxsep=0pt,left=2pt,right=0pt,top=2pt,bottom=2pt,after=\vspace{-5pt} ]
\texttt{<Important Features>}Glucose, Diastolic blood pressure, Fraction inspired oxygen, Respiratory rate, pH\texttt{</Important Features>}
\newline
\texttt{<Final answer>}0.25\texttt{</Final answer>}
\end{tcolorbox}
\\ \hline
\textbf{EAG-RL LLM Output} & 
\textbf{Prediction}: 0.85 → \textbf{\color{blue}{Correct}} \quad | \quad \textbf{Feature Overlap with Expert}: \textbf{3/5}
\newline
\newline
\texttt{<Subquestion 1>}Is the capillary refill rate within normal limits?\texttt{</Subquestion 1>}
\newline
\texttt{<Answer 1>}The capillary refill rate is normal, indicating adequate perfusion.\texttt{</Answer 1>}
\newline
\newline
\texttt{<Subquestion 2>}What is the Glascow Coma Scale (GCS) score, and what does it indicate about the patient's level of consciousness?\texttt{</Subquestion 2>}
\newline
\texttt{<Answer 2>}The GCS score is 15, indicating full consciousness.\texttt{</Answer 2>}
\newline
\newline
\texttt{<Subquestion 3>}Are there any significant changes in blood pressure readings over time?\texttt{</Subquestion 3>}
\newline
\texttt{<Answer 3>}There is a notable drop in diastolic blood pressure from 57.0 to 70.0 mmHg, which may indicate hypotension.\texttt{</Answer 3>}
\newline
\newline
\texttt{<Subquestion 4>}What is the glucose level, and what does it suggest about the patient's metabolic status?\texttt{</Subquestion 4>}
\newline
\texttt{<Answer 4>}The glucose level is very high at 484.0 mg/dL, suggesting uncontrolled diabetes or another cause of hyperglycemia.\texttt{</Answer 4>}
\newline
\newline
\texttt{<Subquestion 5>}How has the patient's heart rate changed over time, and what could this indicate?\texttt{</Subquestion 5>}
\newline
\texttt{<Answer 5>}The heart rate has decreased from 88.0 to 80.0 bpm, which could indicate improvement or the use of beta-blockers.\texttt{</Answer 5>}
\newline
\newline
\texttt{<Subquestion 6>}What is the fraction of inspired oxygen (FiO2) and oxygen saturation, and what do they suggest about the patient's respiratory status?\texttt{</Subquestion 6>}
\newline
\texttt{<Answer 6>}The FiO2 was 100.0\% initially but reduced to 50.0\%, with oxygen saturation remaining stable at 99.0\%, suggesting good oxygenation despite reduced FiO2.\texttt{</Answer 6>}
\newline
\newline
\texttt{<Subquestion 7>}What is the pH level, and what does it indicate about the patient's acid-base status?\texttt{</Subquestion 7>}
\newline
\texttt{<Answer 7>}The pH levels are low at 7.13 and 7.16, indicating metabolic acidosis.\texttt{</Answer 7>}
\newline
\newline
\texttt{<Subquestion 8>}What is the patient's age and sex, and how might these factors influence the prognosis?\texttt{</Subquestion 8>}
\newline
\texttt{<Answer 8>}The patient is an 81-year-old female, factors known to increase mortality risk in ICU settings.\texttt{</Answer 8>}
\newline
\newline
\texttt{<Final subquestion>}Now we can determine the likelihood of the patient not surviving their hospital stay.\texttt{</Final subquestion>}
\begin{tcolorbox}[colback=green!15!white,colframe=white,boxsep=0pt,left=2pt,right=0pt,top=2pt,bottom=2pt,after=\vspace{-5pt} ]
\texttt{<Important Features>}Glucose, Diastolic blood pressure, pH, Age, Sex\texttt{</Important Features>}
\newline
\texttt{<Final answer>}0.85\texttt{</Final answer>}
\end{tcolorbox}
\\ \hline
\end{longtable}
\end{small}
\end{center}
\twocolumn
\end{document}